\documentclass[letterpaper, 10 pt, conference]{ieeeconf}  %

\IEEEoverridecommandlockouts                              

\usepackage{amssymb}
\usepackage{algorithm}
\usepackage{algpseudocode}
\usepackage{graphicx}
\usepackage[flushleft]{threeparttable}
 \usepackage{amsmath} 
\usepackage{subcaption}
\usepackage{multirow}
\usepackage[noadjust]{cite}
\usepackage{xcolor}         
\usepackage[dvipsnames]{xcolor}

\title{Improving Robotic Manipulation Robustness via  NICE Scene Surgery}

\author{
   Sajjad Pakdamansavoji \quad Mozhgan Pourkeshavarz$^{*}$  \quad Adam Sigal$^{*}$ \quad Zhiyuan Li\\ Rui Heng Yang \quad Amir Rasouli
   \thanks{
        Corresponding author \texttt{{sajjad.pakdamansavoji@h-partners.com}}.
         Huawei Technologies Canada.
        $^{*}$Work was done while at Huawei Canada.
        }
}

\begin{document}

\maketitle

\thispagestyle{empty}
\pagestyle{empty}

\begin{abstract}
    Learning robust visuomotor policies for robotic manipulation remains a challenge in real-world settings, where visual distractors can significantly degrade performance and safety. In this work, we propose an effective and scalable framework, \underline{N}aturalistic \underline{I}npainting for \underline{C}ontext \underline{E}nhancement (NICE). Our method minimizes out-of-distribution (OOD) gap in imitation learning by increasing visual diversity through construction of new experiences using existing demonstrations. By utilizing image generative frameworks and large language models, NICE  performs three editing operations, object replacement, restyling, and removal of distracting (non-target) objects. These changes preserve spatial relationships without obstructing target objects and maintain action-label consistency. Unlike previous approaches, NICE requires no additional robot data collection, simulator access, or custom model training, making it readily applicable to existing robotic datasets.
    
    Using real-world scenes, we showcase the capability of our framework in producing photo-realistic scene enhancement. For downstream tasks, we use NICE data to finetune a vision-language model (VLM)  for spatial affordance prediction and a vision-language-action (VLA) policy for object manipulation. Our evaluations show that NICE successfully minimizes OOD gaps, resulting in over 20\% improvement in accuracy for affordance prediction in highly cluttered scenes. For manipulation tasks, success rate increases on average by 11\% when testing in environments populated with distractors in different quantities. Furthermore, we show that our method improves visual robustness, lowering target confusion by 6\%, and enhances safety by reducing collision rate by 7\%.
\end{abstract}


\section{Introduction}
Robustness across visually diverse environments is fundamental for deploying robotic manipulation policies in the real world. Yet, learned policies, especially those trained via behavior cloning on demonstration datasets often suffer from significant performance and safety degradation when presented with visual distractors and scene variations to which they had no exposure during training \cite{pumacay2024colosseum}. To resolve the out-of-domain (OOD) learning gap,  the trivial solution is to collect more data to cover the absent experiences, which can be very time-consuming and resource-intensive. To remedy this problem, some works have explored model-level solutions, such as object-centric representations \cite{chapin2025object, yuan2022sornet} and attention-guided policies \cite{zhang2024lac,james2022coarse}. Others have made use of large-scale simulations, including physics based renderers, domain randomization techniques, and procedurally synthesized scenes, to diversify the training data \cite{robogen_ciml24,zhou2024autonomous,garrett2024skillmimicgen}. However, such solutions are typically dependent on computationally expensive simulators and assume access to large-scale synthetic assets and rendering infrastructure, which are not readily available to practitioners. 

\begin{figure}
    \centering
    \includegraphics[width=1\linewidth]{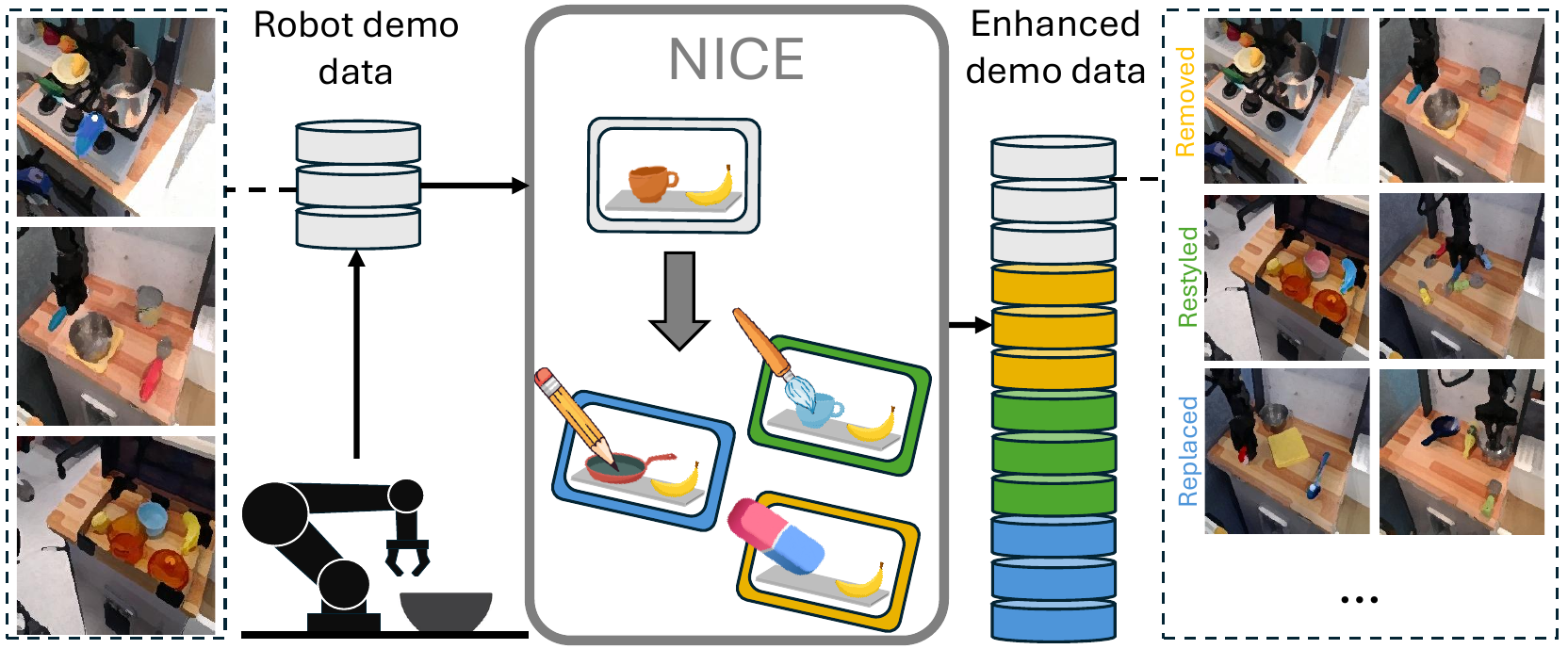}
    \caption{An overview of our NICE generative framework. NICE uses the existing robot demonstration data and performs replacement, restyling, or removal operations on distracting objects to generate new experiences.}
    \label{fig:first_image}
    \vspace{-4mm}
\end{figure}

To this end, we propose  \textbf{\underline{N}}aturalistic \textbf{\underline{I}}npainting for \textbf{\underline{C}}ontext \textbf{\underline{E}}nhancement (NICE). NICE reduces the OOD gap by diversifying demonstration scenes to include varied visual distractors. More specifically, our NICE method applies in\mbox{-}place edits, including replacement with novel objects, restyling with new textures, and removal of distractors along with their shadows. The enhancements are done  directly within real demonstration scenes, keeping the task, viewpoint, and target object fixed. The enhanced data captures scene-level variability common in real-world environments that robots rarely encounter during training due to limited demonstration data. The NICE framework is compatible with any dataset of visual demonstrations and does not require modifications to the underlying robot hardware, control policies, or simulator infrastructure. We conduct evaluations to highlight the realism of our proposed framework, followed by two downstream tasks---visual spatial affordance prediction and object manipulation. We show that the NICE data can mitigate the negative effects of visual distractors on these tasks.
In summary, the contributions of our work is as follow:
\begin{itemize}
    \item We propose NICE, a novel framework for enhancing robotic data with scene-level surgery to improve the robustness of policies to distractors. Our framework effectively scales training data by increasing the variability of contextual elements with minimal human involvement.
    \item We evaluate the realism of NICE data against real-world examples in terms of both background consistency and overall quality of generation.
    \item We highlight the benefit of the NICE data on improving visual affordance prediction in scenes with various levels of clutter caused by distractors.
    \item Via extensive real-world examination, we validate how the NICE data can improve the robustness and safety of robotic manipulation policy across different tasks in environments populated with different numbers of distractors.
\end{itemize}

\section{Related Works}

\textbf{Distractors in visual scene understanding.} In the vision literature, distractors refer to visual elements that are irrelevant to the task at hand yet increase its complexity by diverting attention or introducing ambiguity \cite{liesefeld2024terms}. These distractors may share visual properties with the target object (e.g., color, shape, or saliency), or they may differ in appearance but still act as confounding stimuli.

The impact of distractors has been widely studied across domains. In psychology, numerous studies have investigated how different types of distractors affect visual search \cite{olk2018measuring, petilli2020distractor}, as well as the role of attention mechanisms in mitigating their effects \cite{tsotsos1995modeling, chelazzi2019getting}.Computer vision techniques have also been developed to address distractor-induced challenges, including category-level confusion in object detection \cite{li2021few, liu2022open}, and difficulties in distinguishing targets from visually similar distractors or handling occlusions in tracking tasks \cite{zhu2018distractor, zhong2021towards}.

In robotics, distractors similarly affect performance. For instance, in autonomous driving, a recent work based on the CausalAgents benchmark \cite{sun2024causalagents} shows that modifying irrelevant (non-causal) agents can substantially degrade prediction accuracy, prompting the need for causal reasoning approaches \cite{pourkeshavarz2024cadet, ahmadi2024curb}. For robot localization, the authors of \cite{mendez2021improving} demonstrate that salient distractors can disrupt performance and propose a  technique to suppress their influence. Distractors in cluttered environments are  shown to impact robotic manipulation by interfering with object recognition and complicating grasping actions \cite{dipalo2024kat, kim2020using, kim2021transformer, Samani_2024_persistent, kasaei2024simultaneous, tang2023selective, Ummadisingu2022food}. In some cases, these distractors not only obscure the target but also lead to incorrect action generation. For example, the study in \cite{karamcheti2023voltron} shows that simply altering distractors, either by replacing them with similar items of different color or by swapping them entirely, can reduce the policy's success rate by as much as 50$\%$ across multiple manipulation benchmarks.

\textbf{Visual Robustness in Robotic Manipulation.} Visuomotor policies for robotic manipulation, particularly those trained via behavior cloning, are known to be sensitive to distribution shifts in the visual input space. Prior work has addressed this limitation through architectural enhancements,  such as object-centric representations  \cite{zhu2023groot}, and multimodal foundation models like RT-1 \cite{brohan2022rt1}, which leverage large-scale data and temporal context to improve robustness. Other approaches such as ImitDiff \cite{song2025imitdiff} incorporate semantic segmentation at inference time to guide the policy toward task-relevant regions, demonstrating improved performance under cluttered or distractor-heavy conditions. While these methods improve robustness, they often require complex perception modules or large-scale training infrastructure. Our work instead focuses on improving robustness through data-centric augmentation applied directly to real-world visual demonstrations.

\textbf{Data Augmentation in Robotics.} Domain randomization \cite{tobin2017domain, sadeghi2017cad2rl} has long been used to train visual policies in simulation by exposing models to randomized textures, lighting, and object appearances. However, these techniques are limited to sim-to-real transfer and are rarely applied to real demonstration data. More recently, augmentation methods that directly operate on real robot datasets have shown promise. RoboSaGA \cite{zhuang2025robosaga} employs saliency-guided background replacement using out-of-domain images to preserve task-relevant content while introducing variability. ROSIE \cite{yu2023scaling} uses diffusion models to semantically edit scenes by adding or replacing objects, enhancing generalization to unseen configurations. The method in \cite{chen2024semantically} combines generative image editing with 3D object rendering to generate hundreds of semantically diverse distractor variants per scene. These methods typically rely on large-scale or proprietary generative models or on simulation assets, which may introduce domain gaps or require significant compute, whereas our approach uses an open, reproducible generative component and avoids proprietary dependencies. In addition, they lack studies to showcase their level of realism which can be a source of sim2real gap. In contrast, our method uses direct visual editing to inject distractors into real images with minimal overhead, enabling simple augmentation for existing datasets. We show that our method is effective for generating realistic scenes, resulting in better real-world adaptation.
\begin{figure*}[t]
\centering
\includegraphics[width=0.9\textwidth]{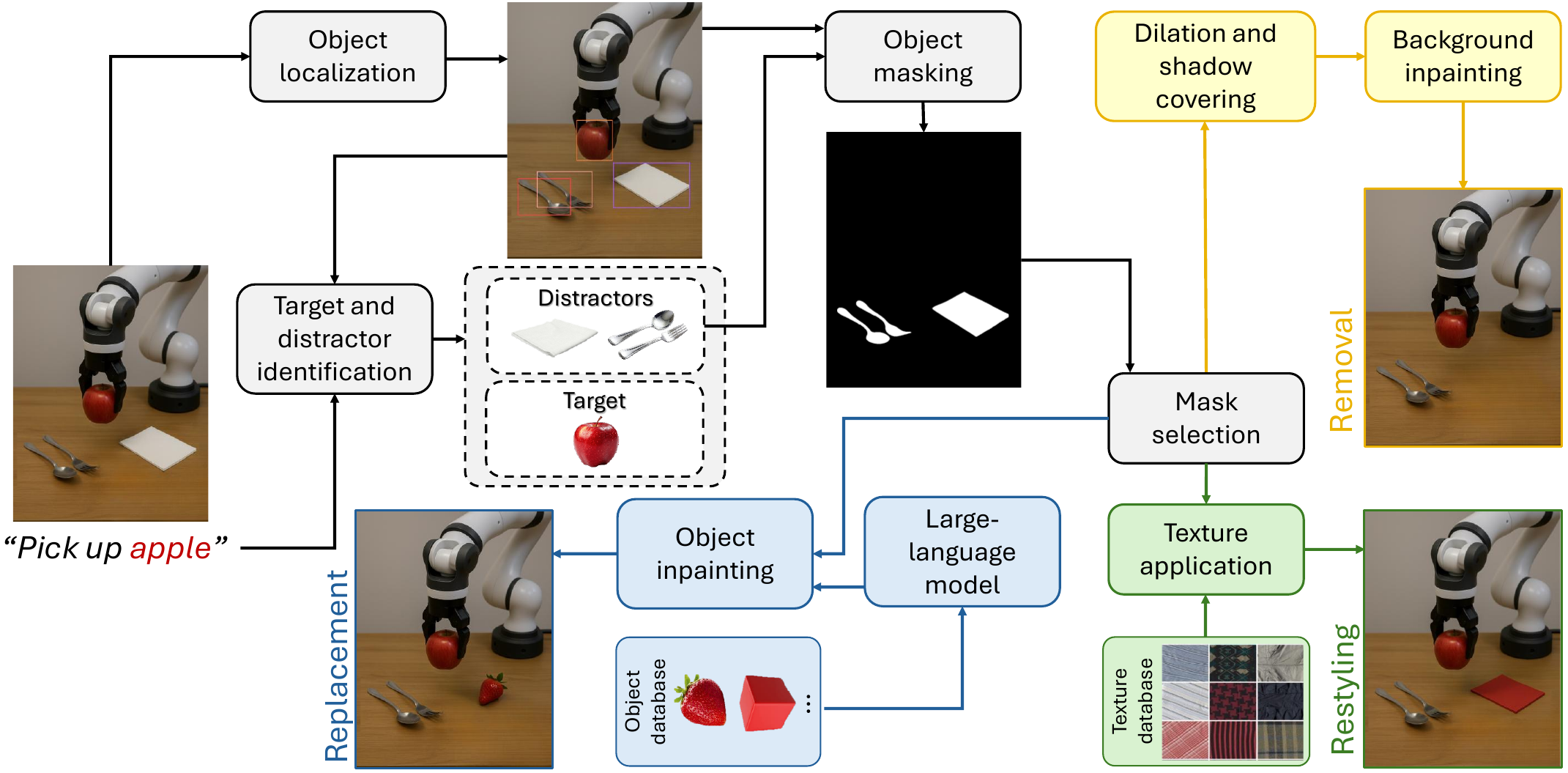}
\caption {An overview of the NICE framework. The method starts by detecting all objects, identifying the target, and segmenting distracting objects. The object of interest is then selected to perform one of the following operations. \textbf{Removal} dilates the selected mask to cover shadows and feeds it into a inpainting model to fill with background texture. \textbf{Restyling} uses a texture database and applies it to the selected mask to change the appearance of the distractor. \textbf{Replacement} uses a large-language model to generate object description, which is then fed into an image inpainting module to replace the distractor.}
\label{fig:overview}
\vspace{-2mm}
\end{figure*}

\textbf{Synthetic Scene Editing.} Beyond robotics, recent works have explored scene editing as a means to evaluate and improve model robustness across a range of domains, underscoring the growing importance of visual generalization. For action recognition, HAT \cite{chung2022enabling} uses video inpainting to isolate or remove humans from real-world clips, revealing strong biases toward background features. In the context of image authenticity detection, Semi-Truths \cite{pal2024semi} introduces localized AI-generated edits into real photographs to test detector sensitivity to subtle manipulations. UltraEdit \cite{zhao2024ultraedit} leverages diffusion-based region editing to build a large-scale benchmark for instruction-based image manipulation. Collectively, these efforts highlight scene alteration as a powerful, domain-agnostic strategy for analyzing and enhancing model robustness. Our work brings this perspective to robotic manipulation, where sensitivity to visual distractors remains a key challenge. By adapting real-world scene editing techniques to robot demonstration data, we contribute both a practical data enhancement pipeline and an evaluation framework grounded in physical robot experiments.

\section{Methodology}

\subsection{Problem Setup}
We consider a standard behavioral cloning setup for visuomotor object manipulation tasks, in which a robot observes RGB images of the scene and outputs corresponding manipulation actions. The training data consist of demonstrations, where each sample includes an image observation, the robot arm’s state, the action executed by the expert policy, and an associated task instruction. The objective is to learn a visuomotor policy that, conditioned on the task instruction, maps observations to actions that imitate the demonstrated behavior. Our aim is to enhance the robustness of policy learning in the presence of visual distractors by enriching the training data with diverse and systematically varied distractor instances while preserving the original task semantics.

\subsection{Overview of NICE}
NICE takes real demonstration images and applies diverse scene enhancements to simulate novel visual distraction, thus generating additional training data. Our framework performs three types of edits: \textbf{removal}, \textbf{replacement}, and \textbf{restyling} of distractors while keeping the original target object and its relation to the demonstration unchanged. A key design principle is action-label consistency, meaning that after enhancement, the image should still correspond to the same grasp or manipulation action as before. To this end, we do not delete or occlude the target object. We further insure that the new instances of the inserted distractors do not conflict with the recorded trajectory. In other words, the task-relevant causal features (e.g. the block to pick up) are invariant under the edits. To preserve the realism of the generated images,  we create separate versions of the same image using one of the three editing operations. As shown in Figure \ref{fig:overview}, the pipeline consists of two stages: Scene decomposition and role assignment, and scene editing.

\subsection{Scene Decomposition and Role Assignment}
\begin{figure}[t]
    \centering
    \includegraphics[width=0.32\linewidth]{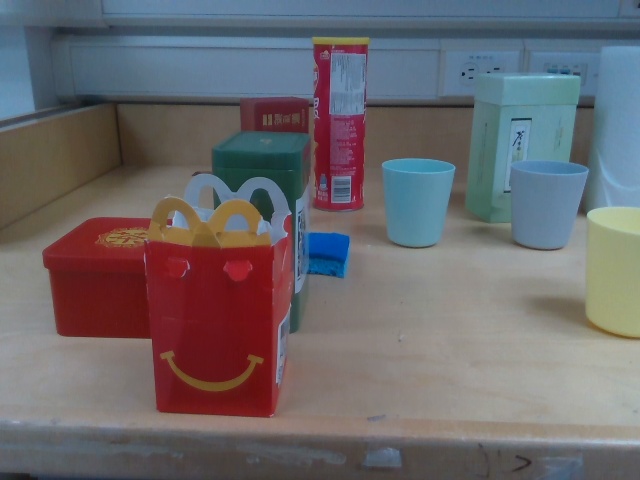}
    \includegraphics[width=0.32\linewidth]{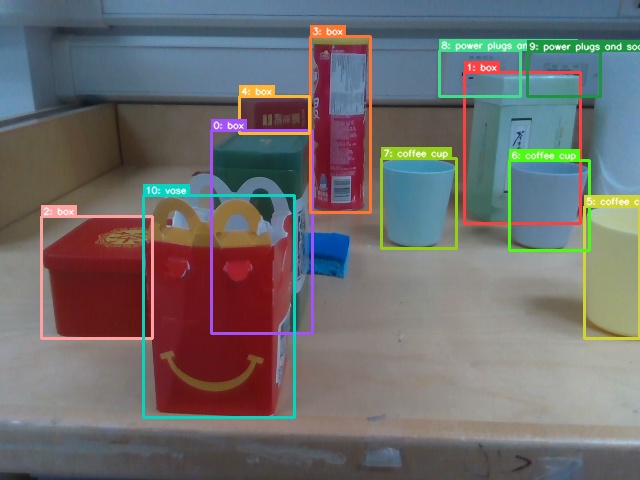} \vspace{0.3em}
    \includegraphics[width=0.32\linewidth]{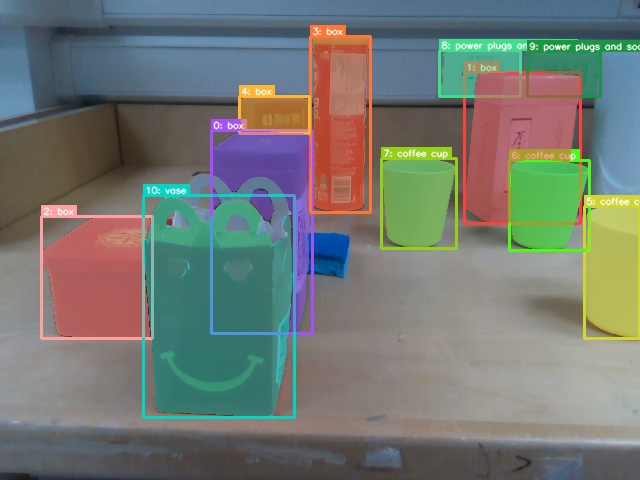}
    \caption{Examples of the object parsing step: (\textbf{Left}) input raw image, (\textbf{Middle}) object detection results using Florence-2, and (\textbf{Right}) segmentation results using SAM-2.}
    \label{fig:fig_det_seg}
    \vspace{-4mm}
\end{figure}

\textbf{Object Parsing.} The first step is to detect all objects in the scene. We use Florence-2 \cite{florence2}, a multitask VLM that detects objects with or without text prompts. Florence-2 produces bounding boxes and class labels for each object. The bounding boxes are then passed to the Segment Anything model v2 (SAM-2) \cite{sam2} to compute precise segmentation masks, along with confidence scores for each object (see Figure \ref{fig:fig_det_seg} for an example).

\begin{figure*}[thbp]
    \centering
    \includegraphics[width=1\linewidth]{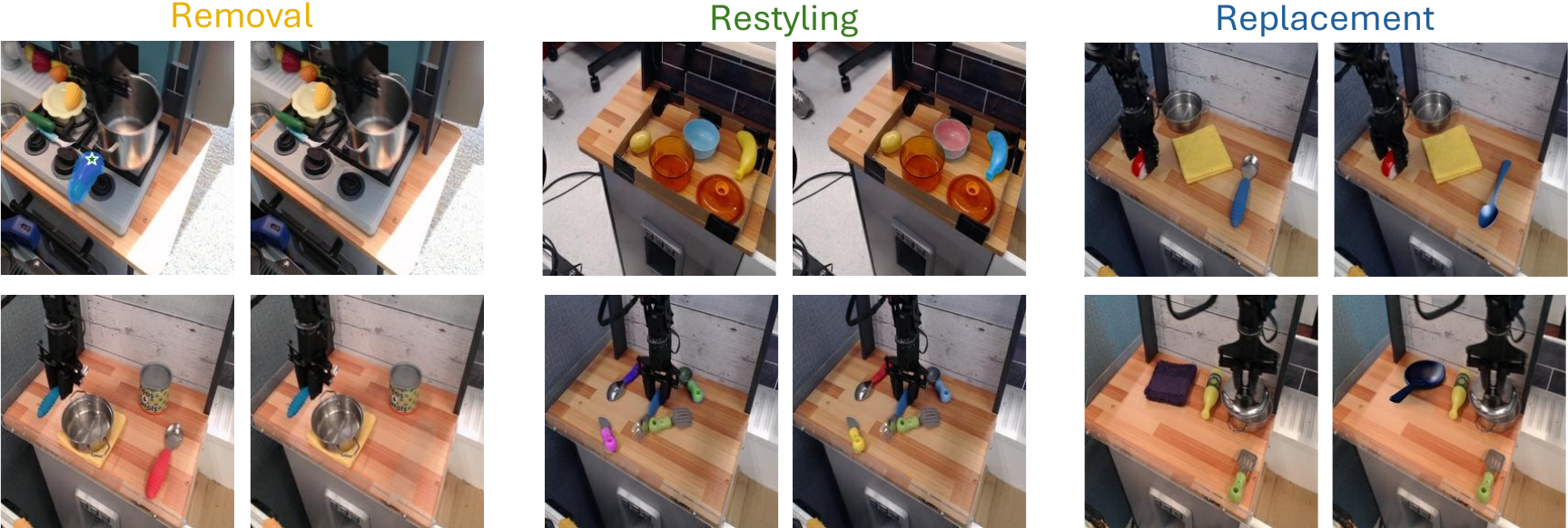}  
    \caption{Examples of data enhancement using NICE on the Bridge data \cite{walke2023bridgedata}. In each image pair, left is the original image and right is the edited one.}
   \label{fig:data_opeartion}
   \vspace{-4mm}
\end{figure*}

\textbf{Target and Distractor Identification.} It is important to accurately distinguish between the target and distractors. Given a task instruction (e.g. pick up the blue cube), we identify the target among all detected objects.  Using the  predicted classes generated by object parsing step, we exclude the target from the segmentation operation. In addition, to improve the consistency of the scenes (e.g. avoid major artifacts in the scene), we exclude very large objects, whose bounding boxes' dimensions exceed 40\% (set empirically) of the image height or width. All other remaining objects are considered as potential candidates for editing.

\subsection{Scene Editing}
For each candidate distractor object, NICE performs one of three edit operations on the copies of the original images (see Figure \ref{fig:data_opeartion} for an example). The operations are performed as follows:

\textbf{Object Removal.} For a given image, a random set of $0$ to $n$ object masks are chosen and combined into a single mask for removal (where $n$ is the number of objects, excluding large size ones or the target). This mask is then dilated with a hyperparameter $dil$ to smooth the edges and cover the original object's shadow. Finally, we mask out the combined distractor region and apply the LaMa inpainting model \cite{lama} to fill it with background content. LaMa is a large-mask image inpainting model based on Fourier convolutions. It propagates texture from surrounding pixels to plausibly reconstruct the scene.

\textbf{Object Restyling.} Our goal is to change the appearance, texture, or color of an object without altering its shape or pose. For this, we follow the same masking strategy as in removal, generating $n$ masks. Then, we sample textures from the Describable Textures Dataset (DTD) \cite{cimpoi14describing}, which contains thousands of real texture patches (e.g. dotted, striped, etc.) applicable to object surfaces. We project the texture onto the object mask by overlaying and adjusting color or by performing stylization. For example, a wooden block might be recolored with a zebra pattern or a metallic spoon with a rust texture. The color and appearance of the objects are altered by adjusting their brightness, hue, and saturation. These transformations are applied to the object masks to introduce controlled variability in visual attributes.

\textbf{Object Replacement.} Unlike object removal and restyling, for each replacement operation, we exchange one object at a time. To maintain realism and consistency, we replace each object with another object that is congruent within the given scene. More specifically, after masking out the target region of the image along with dilation, we use the Stable Diffusion inpainting model \cite{Rombach_2022_CVPR} to generate the recommended object via a structured prompt containing the name of the new object. For example, caption might say ``a yellow block on a wooden table'', and the diffusion model synthesizes the block with appropriate lighting. This insertion leverages state-of-the-art generative priors to produce photo-realistic novel objects.

For replacement, we can employ two different strategies. 1- Generate an instance of the same object category with different appearance, by passing its name to the diffusion model and ask to alter it. 2- Generate a context relevant yet visually distinct object from other categories (e.g. replacing a graphite cooking pan with a gray dish cloth as shown in Figure \ref{fig:data_opeartion}). This allows us to generate a novel scene while maintaining the context. For this purpose, we use Deepseek-r1:7b \cite{guo2025deepseek} via the Ollama framework \cite{marcondes2025using} to generate a description of a household object similar in size to the original one, which is then fed into the Stable Diffusion model \cite{Rombach_2022_CVPR}. In our experiments, we found using such a small language model suffices for accurate prompting in order to generate similar objects.

\section{Evaluation}
\begin{figure}[t]
    \centering
    \includegraphics[width=\linewidth]{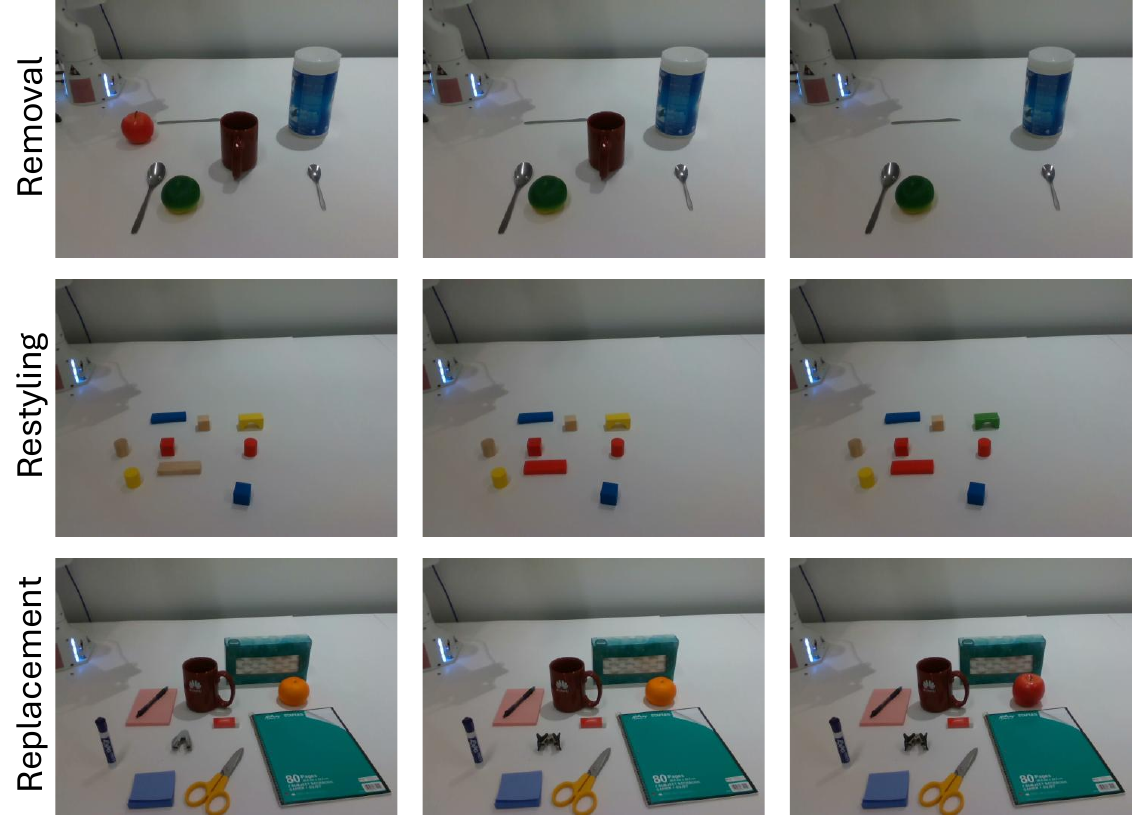}  
    \caption{Real-world replication of editing operations used for validation of the realism of the NICE data. For each series of samples, the scene was populated with multiple objects. Then one at the time, each object was either removed, replaced with the same object with different color, or replaced with another object entirely.}
   \label{fig:real_edit_samples}
   \vspace{-6mm}
\end{figure}

\begin{figure}[t]
    \centering
    \includegraphics[width=0.8\linewidth]{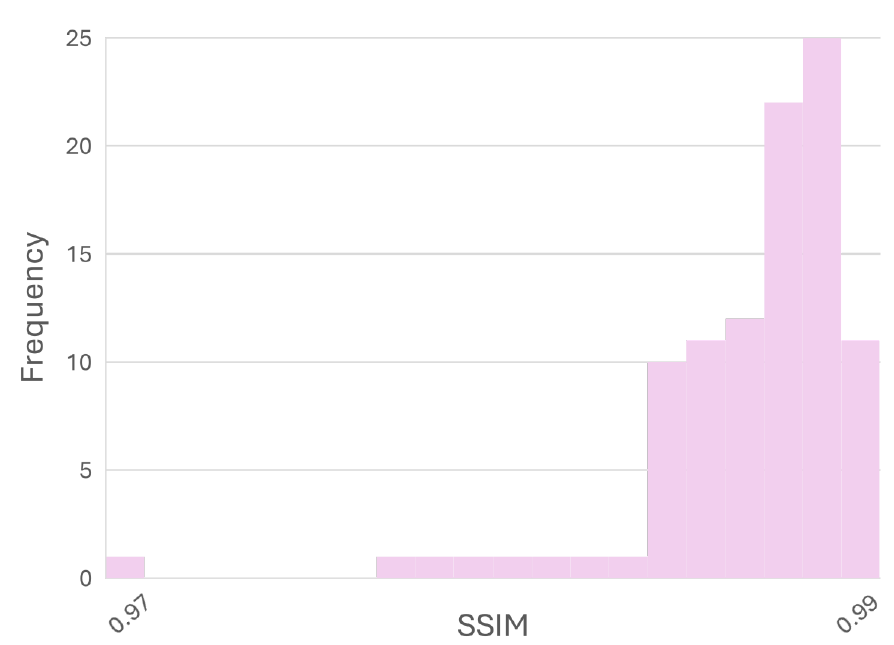}  
    \caption{Distribution of SSIM values for removal operation on real-world data using NICE.}
   \label{fig:ssim_results}
\end{figure}

\subsection{Background Consistency}
\label{subsec:background_consistency}
One of the key considerations for scene editing is to maintain background consistency. This is especially challenging when removing an object, since the background must be reconstructed and secondary effects, such as shadows, must also be eliminated. Here, we examine the ability of our method to achieve this goal in the case of removal. For this, we create 20 cluttered scenes in real world. We then capture 5 variations of the scenes by removing one object at a time, for a total of 100 real-world images (see example in Figure \ref{fig:real_edit_samples}). We then replicate these changes using our pipeline and compare to real images using the SSIM metric \cite{hore2010image}. As shown in Figure \ref{fig:ssim_results}, our method generally yields a very high score on generated samples, indicating its accuracy in reconstructing the background.

\begin{figure}[t]
    \centering
    \includegraphics[width=0.8\linewidth]{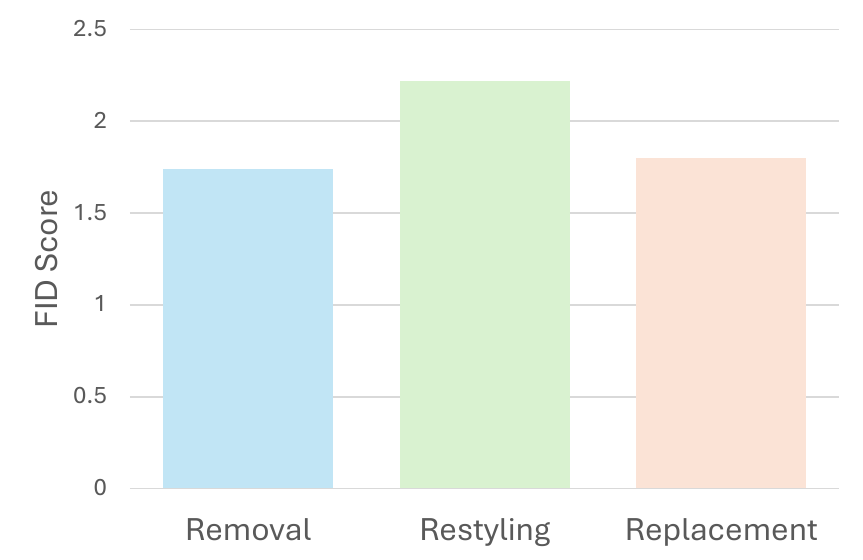}  
    \caption{FID score of the three enhancement strategies on real-world images.}
   \label{fig:fid_results}
   \vspace{-4mm}
\end{figure}

\subsection{Data Generation Realism}
Following the similar procedure as in \ref{subsec:background_consistency}, we capture real-world images for restyling and replacement. For the former operation, we swap the objects with the same objects of different color and for latter, with objects of similar category (e.g. orange with an apple). The real-world data samples are shown in Figure \ref{fig:real_edit_samples}. Using our framework, we then replicate the scene alterations and compute Fréchet Inception Distance (FID) \cite{heusel2017gans} between the generated and real-world captured images. As shown in Figure \ref{fig:fid_results}, lower FID scores indicate that our enhanced images perceptually and statistically are close to the real images. The higher FID value of restyling can be due to the fact that generative models are more successful at modeling ambient conditions (e.g. lighting) when generating an entire object as opposed to restyling the texture of an existing object.

\subsection{Spatial Affordance for Robotics Manipulation}

One of the key issues caused by distractors is visual confusion, which diminishes the ability of the robot to accurately localize the target object and identify affordance regions for performing manipulation. We employ RoboPoint \cite{yuan2024robopoint}, a state-of-the-art vision-language-model (VLM) that predicts spatial affordance in free space, which then can be used for any downstream robotic task.

\begin{figure}
     \centering
        \includegraphics[width=\linewidth]{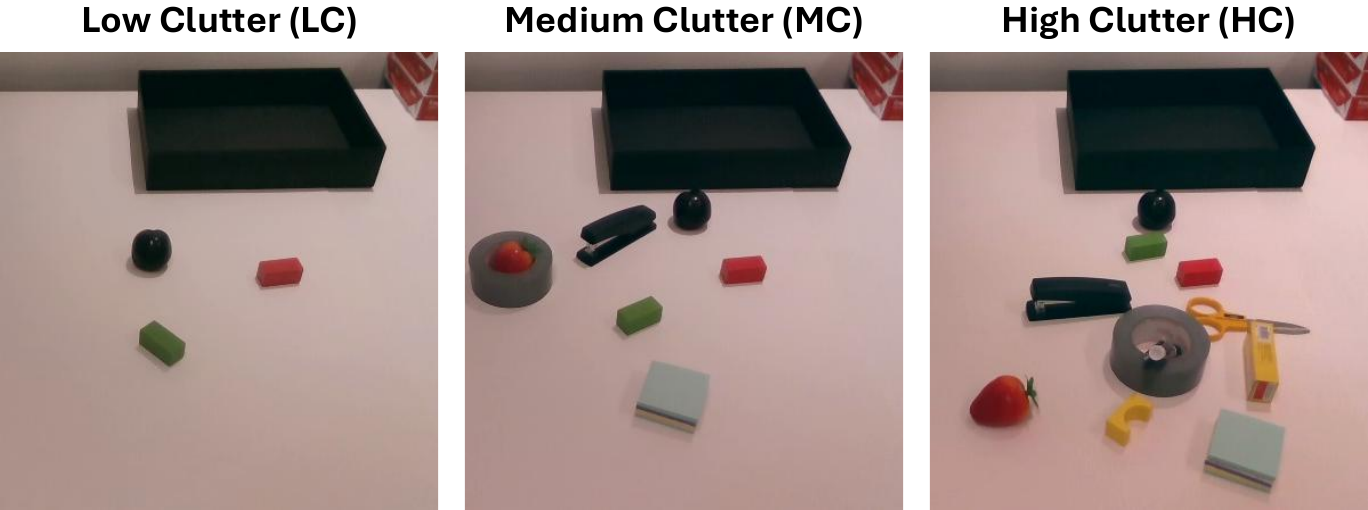} 
        \caption{Samples of scenes with different levels of clutter.}
        \label{fig:fig_cl}
\end{figure}

For this experiment, as shown in Figure \ref{fig:fig_cl}, we consider scenes with three levels of clutter: \textit{low clutter (LC)} with 1-2 objects, \textit{medium clutter (MC)} with 5-8 objects, and \textit{high clutter (HC)} containing 11-15 objects. In every scene, we insert at least one distractor that is visually or semantically similar to the target, as well as additional distractors that differ in category, geometry, or appearance. In high clutter, the objects are densely placed to increase difficulty. Following the protocol in \cite{yuan2024robopoint}, we report the results using \textit{average prediction accuracy (APA)}, which measures the percentage of predicted points that fall within the ground-truth target mask. 

As the results in Table \ref{tab:affordance_res} suggest, our enhancement method can significantly improve the affordance prediction performance. In low and medium clutter scenes we observe an increase of more than 15\% in APA, reaching up to 21\% in high cluttered scenes. This emphasizes the challenge distractors can pose to robot's perception as clutter level increases. Scaling the data using NICE can greatly compensate for such degradation and result in more stable performance across scenes with different levels of clutter.

\begin{table}[t]
 \caption{Average prediction accuracy (APA)$(\%)$ across different clutter levels using RoboPoint \cite{yuan2024robopoint}.} \label{tab:affordance_res}
 \centering
 \resizebox{1\linewidth}{!}{
\begin{tabular}{c|c|c|c}
\hline
\textbf{Dataset} & \textbf{APA$_{\text{LC}}$} & \textbf{APA$_{\text{MC}}$} & \textbf{APA$_{\text{HC}}$} \\ \hline
Original & 32.64 & 30.47 & 20.08 \\ 
\multicolumn{1}{l|}{} & \multicolumn{1}{l|}{} & \multicolumn{1}{l|}{} & \multicolumn{1}{l}{} \\
\begin{tabular}[c]{@{}c@{}} +NICE \end{tabular} & \begin{tabular}[c]{@{}c@{}}\textbf{48.12} \\ \leavevmode\color{Green}(+15.48)\end{tabular} & \begin{tabular}[c]{@{}c@{}}\textbf{45.76} \\ \leavevmode\color{Green}(+15.29)\end{tabular} & \begin{tabular}[c]{@{}c@{}}\textbf{41.44} \\ \leavevmode\color{Green}(+21.36)\end{tabular} \\ \hline
\end{tabular}}
\vspace{-3mm}
\end{table}

\subsection{Robotic Manipulation in Clutter}

\begin{figure*}[t]
    \centering 
    \includegraphics[width=1\linewidth]{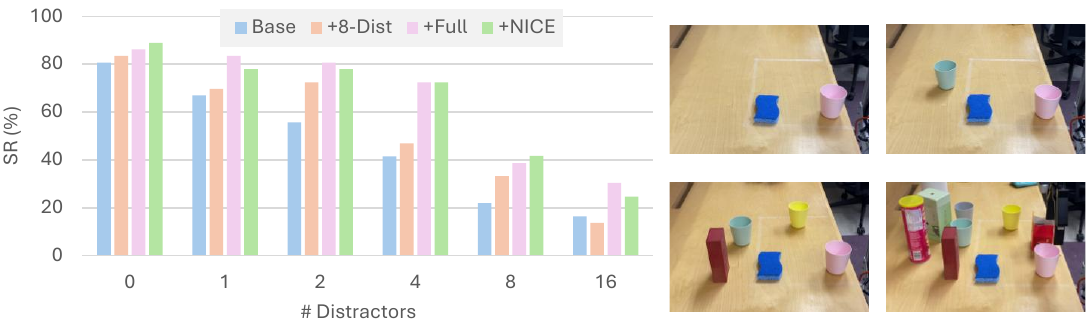}
    \caption{(\textbf{Left}) Performance of the manipulation policy $\pi_0$, finetuned on three data configurations. (\textbf{Right}) Example experiment scenes with varying numbers of distractors.} \vspace{-0.4cm}
    \label{fig:finetuning}
\end{figure*}
In this experiment, we directly measure the impact of the data generated using NICE on the downstream object manipulation tasks. We adopt the four core skills from \cite{li2024evaluating}, namely \textbf{pick} object, \textbf{move} one object close to another, \textbf{put} one object on another, and \textbf{stack} two objects. We design our experimental setup with 6 levels of clutter using 0,1,2,4,8, and 16 distracting objects in the scene. For each setup, we consider 9 different variants (a grid of 3 x 3) for each skill by changing the position of the target object(s). In total, we evaluate on 216 scenarios. Sample scene configurations are shown in Figure \ref{fig:finetuning}.

As our manipulation policy, we choose $\pi_0$ \cite{black2410pi0}, a vision-language-action model pretrained on Open X-Embodiment \cite{open_x_embodiment_rt_x_2023}. We finetune four versions of the policy using the following four datasets; \textbf{Base} data contains 42 demonstrations for each skill collected using only the target objects without any distractors; \textbf{8-Dist} contains data with only 8 distractors with 9 variations for each of the four skills; \textbf{Full} comprises of 45 real demonstrations for each level of clutter for each skill. \textbf{NICE} consists of enhanced data generated using the 8-Dist data as input. For each skill, we create 54 samples by applying 2 variations of each of the 3 enhancement operations. The first model is only finetuned on Base data and subsequent ones are finetuned on Base plus one of the three distracted datasets. We evaluate each finetuned model on all test scenarios, a total of 864 real-world trials.

\textbf{NICE data improves performance at different levels of clutter.}
We begin by measuring the success rate of the policy at different levels of clutter. The results, averaged over all skills,  are shown in Figure \ref{fig:finetuning}. Here, we can see that compared to Base and 8-Dist, NICE data significantly boost the performance of the policy, especially on scenes populated with more distractors. Even though the enhancement is performed on scenes with 8 distractors, the benefit extends to more complex scenes as well, thanks to the diversification of the distractors. Overall, the NICE policy's SR improves upon 8-Dist (which used as input for enhancement) by 11\% and is on par with Full, showing how our method can achieve the same performance without any exposure to manually crafted scenarios.

\textbf{NICE data lowers different types of failures.}
We are interested to determine whether NICE can lower failures, especially those due to visual scene clutter. For this we consider collision rate (CR) measuring the percentage of scenarios in which the arm makes contact with a distractor and target confusion rate (TCR) capturing the percentage of scenarios in which the robot reaches for a non-target object for grasping. As shown in Table \ref{tbl:error_res}, NICE significantly lowers collision rate and target confusion by 7\% and 6\% compared to 8-Dist (which used as seed data for enhancement) and by 12\% and 4\% compared to Full.  Using the same distractors in different settings in manually crafted data, as in 8-Dist, has a positive effect, lowering  CR potentially by 16\%. However, additional samples in the Full dataset results in more overfitting, reducing the advantageous gap. NICE, on the other hand, diversifies the data, leading to improvement of the collision rate even without changing the location of the distractors in the scene. The effect of diversification is also apparent in TCR, allowing the policy to better learn the distinctive features of the target objects. 
\begin{table}[h]
\caption{The average performance of the policy finetuned with each dataset. Abbreviations stand for success rate (SR), collision rate (CR), and target confusion rate (TCR). The direction of arrows indicate higher or lower values are better. } \label{tbl:error_res}
\centering
 \resizebox{0.8\linewidth}{!}{
\begin{tabular}{l|c|c|c}
\hline
                 & \textbf{SR$\uparrow$}   & \textbf{CR$\downarrow$}   & \textbf{TCR$\downarrow$}    \\ \hline
\textbf{Base}    & 47          & 38          & 34                   \\
\textbf{+8-Dist} & 53          & 22          & 16                   \\
\textbf{+Full}   & \textbf{65}         & 27          & 14                    \\ \hline
\textbf{+NICE}   & 64 & \textbf{15} & \textbf{10} \\ \hline
\end{tabular}}
\end{table}

 Despite lowering CR and TCR significantly, NICE achieves similar SR to Full. There are two main reasons for this: 1- as per our evaluation, we don't consider collision as a failure as long as the task is completed. Hence, NICE achieves similar level of success but more safely. 2- There are other types of failures, e.g. grasp failure, placement failure, etc. which can contribute to lack of success. For instance, since in NICE we do not alter the position of the target objects, their poses are not diversified, hence no added advantage is achieved. The Full data, on the other hand, contains scenes with different setups for the targets, leading to more robust grasping. 

\textbf{NICE benefit amplifies as the tasks get more complex} We consider the per-task breakdown of the results in Table \ref{tbl:skill_perf}. Here, we can see that the more complex the tasks get, the more benefit is gained using the NICE data. In the Pick task, which only involves lifting the target, the NICE policy lags behind in SR due to lower advantage in improving grasp robustness (as discussed in the previous experiment). However, it still drastically lowers collision rate. The reduction in CR extends to more complex tasks, where the gap with Full becomes even higher, 28\% in Put and 12\% in Stack tasks. The reason for this is that in these tasks two targets are involved, the object that is to be grasps and the one on which the target is to be placed on. Hence manipulating one with respect to the other, increases the likelihood of collision with distractors. Overall, compared to the Base data and the 8-Dist data, NICE clearly stands out on all metrics for all tasks.

\begin{table}[t]
\caption{The performance of the policy average per skill. Results are shown as success rate (SR)/collision rate (CR). For the former higher is better and latter lower.}\label{tbl:skill_perf}
\centering
 \resizebox{1\linewidth}{!}{
\begin{tabular}{l|cccc|c}
\hline
                 & \textbf{Pick} & \textbf{Move} & \textbf{Put} & \textbf{Stack} & \textbf{Average} \\ \hline
\textbf{Base}    & 48/46         & 52/30         & 44/39        & 45/37          & 47/38            \\
\textbf{+8-Dist} & 59/13         & 56/28         & 50/37        & 48/11          & 53/22            \\
\textbf{+Full}   & \textbf{80}/20         & 65/19         & 58/54        & \textbf{59}/17          & \textbf{65}/27            \\ \hline
\textbf{+NICE}   & 67/\textbf{9} & \textbf{67/20}  & \textbf{63/26} & \textbf{59/5}   & 64/\textbf{15}    \\ \hline
\end{tabular}}
\vspace{-3mm}
\end{table}





\section{Conclusion and Future Work}
In this work, we proposed a novel approach for enhancing robot data without the need for action generation or human involvement. Our NICE method, relies on a language conditioned generative model to identify the objects of interest and performs scene editing by either removing, restyling, or replacing distractors with novel objects or background. Through empirical evaluation on real-world data, we showed that our pipeline generates realistic scenes that significantly improve robot perception and, consequently, downstream manipulation tasks.

In this work we mainly focused on three forms of scene enhancement, namely removal, restyling, and replacement. We argued that correct selection of novel objects can maintain the realism of the scenarios, e.g. not obscuring robot movement in the pre-recorded scenes. For data generation, other forms of enhancement can be considered, such as rearrangement or addition. However, these operations require better understanding of robot actions in the 3D space to maintain realism. We will consider such extensions for our future work. 

In this work, we examined the impact of our framework on spatial affordance prediction and core manipulation tasks. It is reasonable to assume that visual confusion or operational confinement caused by distractors can have different degrees of impact on different manipulation tasks. For example, relative to object-picking tasks, object arrangement poses greater challenges, as it increases the likelihood of confusion. We plan to extend our empirical evaluation on a large set of robotic skills to both identify challenges posed by distractors and clutter and determine whether our proposed data enhancement framework can be used to mitigate them.
\bibliographystyle{IEEEtran}

\bibliography{ref}

\end{document}